\documentclass[11pt]{article}
\usepackage[]{acl}
\usepackage{latexsym}
\usepackage[T1]{fontenc}
\usepackage[utf8]{inputenc}
\usepackage{booktabs}
\usepackage{graphicx}
\usepackage{amsmath}
\usepackage{amssymb}
\PassOptionsToPackage{hyphens}{url}
\usepackage{url}
\title{When a Data Artifact Isn't a Shortcut: Causal Auditing of
       Synthetic RLVR Corpora}

\author{
  Esther Xin \\
  Independent Researcher \\
  \texttt{estherxin0011@gmail.com}\\
  {\small Code and data: \url{https://github.com/ethxin0011/rlvr_authenticity_audit}}
}

\begin{document}
\maketitle

\begin{abstract}
Several recent pipelines build RLVR training data by masking a span of
real corpus text and asking a language model to invent plausible wrong
answers around it. The correct option is therefore genuine human prose;
every distractor is synthetic. Correctness and provenance become
entangled, and a policy could in principle learn the second instead of
the first. We audit that possibility in GooseReason-0.7M. First we ask
whether the asymmetry is visible at all: a classifier reading only five
surface statistics---never the meaning---reaches AUROC 0.562 over
315{,}499 options, barely above chance. The aggregate hides something,
though. Code sits at 0.416, \emph{below} chance, and manual inspection
explains why: code distractors turn out to be single-operator mutations
of the gold answer rather than freely written alternatives, so the two
classes are nearly identical by construction. Detecting a signal is not
the same as showing a model uses it, so we then run an intervention.
We build a paraphrase-matched control corpus, hold training-set size
identical across arms, and train two policies under one fixed budget.
The exploitation gap does not favour the unmodified-data arm: 0.021
against 0.027 for the control. Under our budget, in other words, a
detectable artifact went unexploited. We think that dissociation, along
with the domain-specific construction finding, is worth knowing for
anyone curating corpora of this kind, and we release the audit as a
mostly CPU-only protocol.
\end{abstract}

\section{Introduction}
\label{sec:intro}

RLVR has become a default recipe for improving reasoning
\citep{guo2025deepseekr1}, but it consumes verifiable data faster than
anyone can curate it. One response is to manufacture that data.
\citet{lu2026goldengoose} mask a reasoning-critical span in a source
document, prompt a model for plausible but wrong replacements, and
recover a multiple-choice item whose answer a program can check. Applied
at scale, the recipe yields GooseReason-0.7M: 700{,}000 items spanning
mathematics, code, and general STEM.

Something about this construction has gone unexamined. The correct
option is real text, lifted from a real document. Every distractor was
written by a model. So the option classes differ twice over---in
correctness, which is the signal the task intends, and in provenance,
which is not. Should provenance leave any statistical residue, in token
rarity or sentence rhythm or syntactic shape, a policy might learn to
read that residue and skip the reasoning entirely.

This is not the failure mode the literature has already covered.
\citet{guo2026rethinkingmcq} show policies eliminating weak distractors
instead of reasoning through them; \citet{tang2026positionbias} show
generated items carry answer-\emph{position} bias. Both concern the
distractor set---how strong it is, where the answer sits. Neither
touches the provenance of the text itself. And neither asks the question
we think matters most: granted an asymmetry exists, does a trained
policy actually use it?

We separate those two questions deliberately, because conflating them is
easy and misleading. \textbf{Stage 1} characterizes: can surface
statistics alone tell real spans from generated ones, at scale, with no
access to meaning? \textbf{Stage 2} intervenes: we construct a
paraphrase-matched control corpus that removes the asymmetry while
holding difficulty and training-set size fixed, train matched policies
on original and control data under one budget, and test whether either
leans harder on authenticity-correlated cues when probed adversarially.

\paragraph{Findings.} A surface-only classifier lands near chance
overall (AUROC 0.562, $n=315{,}499$), and math and STEM are similar
(0.584, 0.583). Code breaks the pattern at 0.416---below chance---which
manual audit attributes to a different generation mechanism entirely:
its distractors are minimal mutations of the gold code, not independent
rewrites. Turning to Stage 2, the policy trained on unmodified data
shows no larger exploitation gap than the control-trained one (0.021
vs.\ 0.027, sizes matched). Two of three domains hold that ordering.
Math shows a small \emph{negative} gap---and shows it in both arms,
which is why we read it as a domain property rather than noise in one
run.

\paragraph{Contributions.} We offer (i) a cheap, semantics-blind
diagnostic for provenance asymmetry in corpora that mix real and
generated text, runnable before any GPU budget is committed; (ii) the
first audit we know of into GooseReason-0.7M's authenticity structure,
which surfaces an undocumented per-domain difference in how distractors
are made; (iii) a causal protocol---paraphrase-matched controls with
training-set size pinned across arms---for testing whether a detected
asymmetry is actually exploited; and (iv) evidence that detection and
exploitation can come apart, since a real if weak signal did not
translate into measurably greater shortcut reliance at our budget.

\section{Related Work}
\label{sec:related}

\paragraph{Synthesizing RLVR data from unverifiable text.}
The masked-span-plus-distractor construction we audit comes from
\citet{lu2026goldengoose}, who report strong downstream gains from
training on the result. Their evaluation measures downstream accuracy,
as most papers introducing a synthesis pipeline do; it does not
characterize the statistical shape of the distractors themselves. We see
our audit as orthogonal rather than adversarial. A dataset can lift
downstream accuracy and still carry a construction asymmetry worth
documenting before it scales further.

\paragraph{Distractor design and position bias.}
\citet{guo2026rethinkingmcq} demonstrate that weak distractors invite
elimination-based shortcuts, and propose iterative curation to harden
against them. \citet{tang2026positionbias} document systematic,
model-family-specific position bias in generated MCQs. Each targets a
property of the distractor set as a whole---discriminative strength in
one case, positional placement in the other. Our axis is different: not
where the answer sits or how strong any single distractor is, but
whether the option text carries a real-versus-generated signature at
all.

\paragraph{Provenance detection.}
Our Stage 1 features---rare-token rate, burstiness, entropy---come
recognizably from the machine-generated-text detection toolkit, and we
use them deliberately. The purpose differs, though.
\citet{wu2025syntheticaudit} audit deployed artifacts to establish
whether synthetic data was used in training them, a forensic question
about a model's history. We are not attributing authorship. We are
asking whether an RLVR corpus's construction leaves a trace strong
enough to separate its option classes, and then whether that trace
matters downstream.

\paragraph{Shortcut learning.}
That models exploit dataset artifacts rather than solving tasks is well
established \citep{du2023shortcut}. What is less examined, and what we
target, is the converse case: an artifact present and measurable, yet
apparently unused.

\paragraph{Verifier and reward reliability.}
A parallel literature studies unreliability in the \emph{reward} rather
than the \emph{data}. \citet{cai2025noisy} treat verifier error as a
stochastic reward channel and derive policy-gradient corrections.
\citet{egashira2026delay} find that whether verifier errors are
systematic or random governs the outcome, with systematic error capable
of producing plateaus or collapse regardless of aggregate error rate.
\citet{huang2025accuracy} show rule-based and model-based verifiers fail
in different directions; \citet{helff2026gaming} show policies learning
to game an imperfect verifier's blind spots; \citet{ray2026fuzzing}
fuzz verifiers directly to expose those spots pre-training; and
\citet{norman2026reliability} document a reliability-without-validity
gap in LLM-as-judge evaluators at scale. The common lesson is that a
detectable flaw in an RLVR component does not automatically imply it is
exploited, or exploited evenly---precisely the systematic-versus-random
distinction \citet{egashira2026delay} draw. We borrow that logic and
move it upstream, from the verifier to the training corpus. A detectable
authenticity asymmetry in the data need not be an exploited one, and
Stage 2 is, to our knowledge, the first direct causal test of that
proposition for synthetic RLVR data construction.

\section{Method}
\label{sec:method}

\subsection{Data and construction mechanism}
\label{sec:data}
GooseReason-0.7M \citep{lu2026goldengoose} covers mathematics, code, and
general STEM reasoning. Each item begins as a source passage with a
reasoning-critical span masked out; a language model then supplies
plausible-but-wrong alternatives for that span. Restore the masked span
and it becomes the correct option. Everything else on the page was
generated. That is the asymmetry we audit: gold provenance is the
original corpus, distractor provenance is the generator, and this holds
regardless of domain.

\subsection{Stage 1: surface-feature characterization}
\label{sec:stage1}
For every option---gold and distractor alike---we compute five features
that require no interpretation of meaning: rare-token rate (share of
tokens below a Zipf-frequency threshold), word- and sentence-length
burstiness (variance-to-mean ratio of per-token and per-sentence
character length), character-level Shannon entropy, and POS-tag
KL-divergence against a fixed reference distribution. Two classifiers,
logistic regression and gradient-boosted trees, are trained to predict
provenance from those five numbers alone, on a stratified 75/25 held-out
split; we report AUROC overall and per domain. Because nothing in the
feature set reads content, any classifier that beats chance has found a
genuine authenticity signal, whatever its size.

\subsection{Stage 2: causal intervention}
\label{sec:stage2}
Stage 1 can establish presence. It cannot establish use. For that we
build a paraphrase-matched control corpus in which every option---gold
and distractor alike---passes through the same transformation, so the
authenticity asymmetry is removed while option order, and therefore the
correct letter, survives intact.

The transformation is domain-conditioned, and the reason is worth
stating. For math and STEM we apply full semantic paraphrase. For code
we do not. A manual audit of ten sampled code items found, in every
single one, that the distractor was a one-token mutation of the gold
code---a comparison flipped from \texttt{<} to \texttt{>}, a loop bound
moved by one---rather than an independently composed alternative. Run a
generic paraphraser over that and it may quietly repair the injected
fault, which would destroy the only thing making the item solvable. So
code instead receives a restyle-only transform: rename variables,
adjust formatting and comments, under explicit instruction to leave
control flow, operators, and correctness untouched. Stylistic
authenticity cues go; the injected fault, and the difficulty it
carries, stays.

Two policies then train under otherwise identical settings---a
\emph{treatment} policy on original data, a \emph{control} policy on the
paraphrase-matched version. Per-domain training-set size is capped
identically for both (Table~\ref{tab:hyperparams}). This matters more
than it might appear: build the control corpus independently of the
treatment corpus's sampling and it is easy to end up comparing a
data-authenticity effect against a data-quantity effect without
noticing.

\subsection{Evaluation protocol}
\label{sec:eval}
Both policies face a held-out test partition, unseen in training, in
three variants built from the same underlying items. \textbf{Original}
keeps the data unmodified, asymmetry present. \textbf{Neutralized} uses
the paraphrase-matched version, asymmetry removed. \textbf{Adversarial}
pairs the neutralized gold option with one unmodified,
authenticity-bearing distractor substituted back in. A policy leaning on
an authenticity cue should be pulled toward that authentic-looking wrong
option, and the pull should show up as an accuracy drop relative to the
neutralized variant. We therefore define the \textbf{artifact
exploitation gap} as
\[
  \text{gap} = \text{acc(neutralized)} - \text{acc(adversarial)},
\]
reported per policy, overall and by domain. A larger gap in the
treatment arm is the signature of the hypothesis; a gap that is equal or
smaller is evidence against it, at the budget tested.

\subsection{Reward, training, and compute}
\label{sec:training}
Both policies fine-tune from Qwen3-1.7B with GRPO
\citep{shao2024deepseekmath} over 4-bit quantized low-rank adapters. The
reward is rule-based and fully verifiable---an exact match between the
emitted option letter and the correct one---so no learned verifier's
biases leak into the causal comparison.
Table~\ref{tab:hyperparams} gives the full configuration.
\S\ref{sec:limitations} states plainly that this budget is modest by
current standards, and what that costs us in strength of claim.

\section{Results}
\label{sec:results}
\begin{table}[htbp]
\centering
\small
\setlength{\tabcolsep}{6pt}
\begin{tabular}{lrrr}
\toprule
Split & GBM AUROC & LogReg AUROC & $n$ \\
\midrule
overall & 0.562 & 0.551 & 315{,}499 \\
math    & 0.584 & 0.598 & 101{,}926 \\
stem    & 0.583 & 0.569 & 102{,}414 \\
code    & \textbf{0.416} & 0.503 & 111{,}159 \\
\bottomrule
\end{tabular}
\caption{Held-out AUROC for a classifier separating gold spans from
distractors using five surface-statistical features alone (rare-token
rate, word- and sentence-length burstiness, character entropy, and
POS-tag KL-divergence against a reference distribution). No feature
inspects semantic content. \emph{code} is the only split below the
chance line of 0.500.}
\label{tab:phase1auroc}
\end{table}

\begin{table}[htbp]
\centering
\small
\setlength{\tabcolsep}{5pt}
\begin{tabular}{lrrrr}
\toprule
Group & Orig. & Neutr. & Advers. & Gap \\
\midrule
treatment & 0.283 & 0.212 & 0.191 & 0.021 \\
control   & 0.299 & 0.233 & 0.207 & 0.027 \\
\bottomrule
\end{tabular}
\caption{Overall four-option MCQ accuracy (chance $=0.25$) for the
treatment-trained (original data) and control-trained (paraphrase-matched
data) policies on three held-out test variants, and the resulting
artifact exploitation gap (Neutr.\ $-$ Advers.). The control gap is not
smaller than the treatment gap, giving no evidence that training on
unmodified data induces additional shortcut reliance relative to training
on artifact-neutralised data.}
\label{tab:phase2overall}
\end{table}

\begin{table}[htbp]
\centering
\small
\setlength{\tabcolsep}{4pt}
\begin{tabular}{llrrrr}
\toprule
Group & Domain & Orig. & Neutr. & Advers. & Gap \\
\midrule
treatment & code & 0.300 & 0.220 & 0.187 & 0.033 \\
treatment & math & 0.153 & 0.103 & 0.137 & $-$0.033 \\
treatment & stem & 0.397 & 0.313 & 0.250 & 0.063 \\
control   & code & 0.330 & 0.263 & 0.207 & 0.057 \\
control   & math & 0.167 & 0.123 & 0.150 & $-$0.027 \\
control   & stem & 0.400 & 0.313 & 0.263 & 0.050 \\
\bottomrule
\end{tabular}
\caption{Exploitation gap decomposed by domain. The negative gap in
\emph{math} is not an artefact of a single arm: it recurs, at similar
magnitude, in both the treatment and control policies, indicating a
domain-level effect independent of training-data provenance rather than
sampling noise in one arm.}
\label{tab:phase2domain}
\end{table}

\subsection{A weak signal, unevenly distributed}
\begin{figure}[htbp]
\centering
\includegraphics[width=\columnwidth]{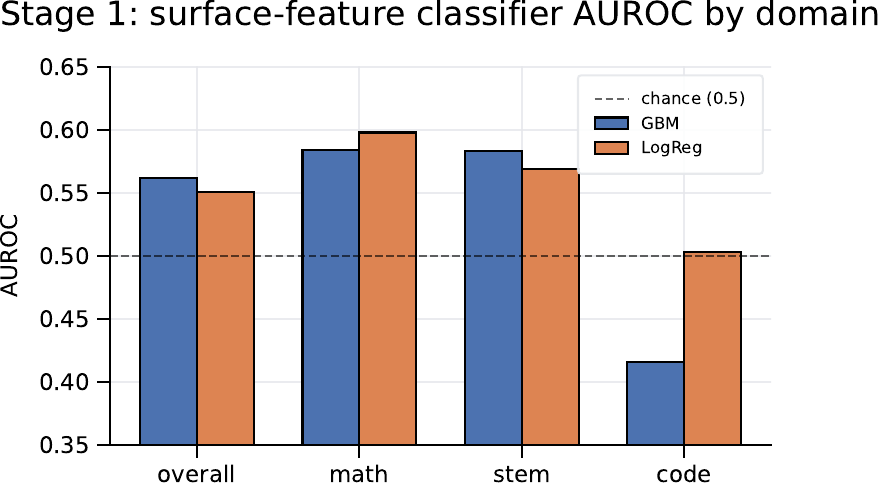}
\caption{Stage 1 classifier AUROC by domain and classifier type. The
dashed line marks chance (0.5). \emph{code} is the only domain falling
below it.}
\label{fig:stage1}
\end{figure}

Table~\ref{tab:phase1auroc} reports Stage 1. Overall AUROC sits at
0.562, close enough to chance to be unremarkable, and math and STEM are
much the same (0.584, 0.583). Code is the outlier, at 0.416---below
chance. Figure~\ref{fig:stage1} shows the spread.

A below-chance AUROC on held-out data is not a stronger signal running
backwards. It indicates that surface statistics are, if anything,
mildly \emph{anti}-informative here under our feature set. The manual
audit explains it: when distractors are one-token mutations of the gold
answer, the two classes are nearly identical in surface statistics by
construction, and a surface-only classifier has little to learn in
either direction.

\subsection{No exploitation under RLVR training}
\begin{figure}[htbp]
\centering
\includegraphics[width=\columnwidth]{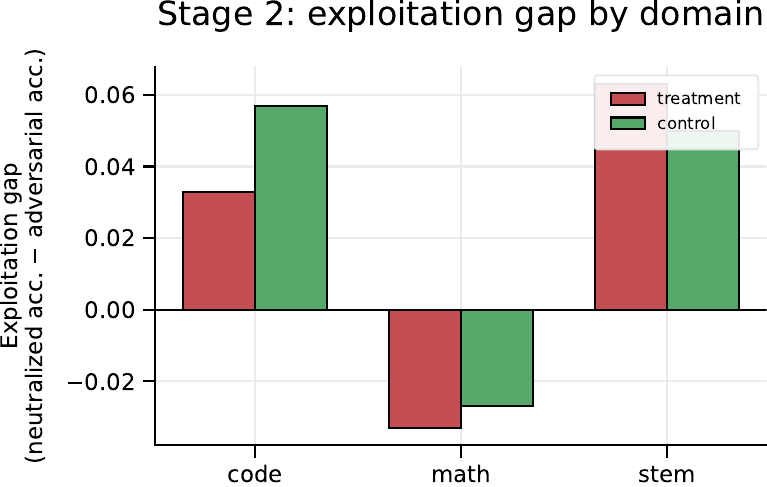}
\caption{Exploitation gap by domain, treatment vs.\ control policy. No
domain shows a treatment gap substantially exceeding its control
counterpart.}
\label{fig:stage2}
\end{figure}

Table~\ref{tab:phase2overall} gives the overall gap. Treatment reaches
0.021, control 0.027---so the treatment gap is not larger, and the
ordering, if anything, runs the other way.
Table~\ref{tab:phase2domain} and Figure~\ref{fig:stage2} break this
down. In code and STEM the control gap matches or exceeds treatment's.
In math both policies produce a small \emph{negative} gap, higher
accuracy on adversarial than neutralized, at close magnitudes
($-0.033$ treatment, $-0.027$ control). Because the pattern appears in
two independently trained policies rather than one, we attribute it to
the math domain's adversarial construction under this evaluation design
and not to noise localized in a single run.

Put together, the two stages answer two questions and give two different
answers. An authenticity asymmetry is present in the data, weak and
unevenly spread. Under our budget it did not translate into measurably
greater reliance on authenticity-correlated shortcuts, comparing a
policy trained on unmodified data against one trained on neutralized
data at matched difficulty and matched size.

\section{Discussion}
\label{sec:discussion}

\paragraph{Detectable does not imply exploited.} The central result is a
dissociation. A classifier finds a domain-specific, statistically real
authenticity signal in Stage 1; that signal does not show up as a
measurable difference in shortcut reliance in Stage 2. The shape of this
mirrors, one level upstream, the distinction
\citet{egashira2026delay} draw about verifier error---that whether a
flaw matters depends on more than whether it can be detected in
isolation. Our suggestion for future audits of synthetic RLVR corpora:
where compute permits, pair cheap characterization with a causal
intervention before concluding a detected asymmetry needs engineering
around. Answering only the first question risks mis-stating the stakes
in either direction.

\paragraph{Report the mechanism, not just the output.} The code finding
was not visible in the dataset's description of a single uniform
synthesis procedure. It emerged only by combining a below-chance
classifier result with hands-on inspection of individual items. Datasets
built by prompting a model for distractors would be more auditable if
they reported, per domain, whether the generation mechanism is actually
uniform. As here, a mechanism that varies by domain can produce a
qualitatively different failure surface---bug-injection mutation versus
free rewriting---that no single aggregate statistic will reveal.

\paragraph{Relation to distractor-strength and position-bias work.}
\citet{guo2026rethinkingmcq} find policies exploiting \emph{weak}
distractors through elimination; \citet{tang2026positionbias} find
positional bias in generated items. Nothing here contradicts either. A
policy might well lean on distractor weakness or position while showing
no authenticity-shortcut gap at all, since these are independent axes of
the same synthetic data. Our protocol complements those audits rather
than replacing them, each aimed at a different candidate shortcut in the
same family of pipelines.

\section{Limitations}
\label{sec:limitations}
\textbf{Modest RLVR budget.} Sixty GRPO steps on one 16\,GB GPU
(Table~\ref{tab:hyperparams}) sits well below production RLVR scale. A
null exploitation gap here argues against a \emph{readily} learned
shortcut. It does not establish that none would emerge with far longer
training, and we report the budget explicitly rather than leave that
scope implied. \textbf{Single seed per arm.} Each policy trained once,
so we have no variance estimate across runs; the small effects we report
should be read with that in mind. \textbf{Per-domain sample size.} Three
hundred test items per domain, 900 total---enough for the differential
gap our design targets, not enough to pin down very small per-domain
effects. \textbf{One corpus, one construction family.} We audit a single
dataset from one masked-span-and-distractor procedure. Whether the
dissociation generalizes to other real-versus-generated construction
methods is open. \textbf{MCQ format only.} Our reward and evaluation
target multiple-choice RLVR. The same asymmetry could behave differently,
or not arise, in free-form or code-execution settings.

\section{Conclusion}
\label{sec:conclusion}
Pairing real corpus text as ground truth with model-generated
distractors introduces an authenticity asymmetry alongside the
correctness signal a task intends. We audited that asymmetry in
GooseReason-0.7M across two stages, built to keep apart two questions
that are easy to run together: is an asymmetry detectable, and is it
exploited? We got both answers, and they diverge. A surface-detectable
signal exists, driven on inspection by a distinct code-domain mechanism
producing near-identical bug-injection distractors rather than free
rewrites. Yet training an RLVR policy on unmodified data did not
measurably raise shortcut reliance against an artifact-neutralized
control at matched difficulty and matched training-set size. We release
the characterization protocol, the paraphrase-matched control
construction, and the per-domain evaluation splits, so the audit can be
pointed at other synthetic RLVR corpora, and repeated at budgets larger
than ours.

\bibliography{references}

@article{guo2025deepseekr1,
  title   = {{DeepSeek-R1}: Incentivizing Reasoning Capability in {LLM}s
             via Reinforcement Learning},
  author  = {Guo, Daya and Yang, Dejian and Zhang, Haowei and others},
  journal = {arXiv preprint arXiv:2501.12948},
  year    = {2025}
}

@article{shao2024deepseekmath,
  title   = {{DeepSeekMath}: Pushing the Limits of Mathematical Reasoning
             in Open Language Models},
  author  = {Shao, Zhihong and Wang, Peiyi and Zhu, Qihao and Xu, Runxin
             and Song, Junxiao and Bi, Xiao and Zhang, Haowei and
             Zhang, Mingchuan and Li, Y. K. and Wu, Y. and Guo, Daya},
  journal = {arXiv preprint arXiv:2402.03300},
  year    = {2024}
}

@article{lu2026goldengoose,
  title   = {Golden Goose: A Simple Trick to Synthesize Unlimited {RLVR}
             Tasks from Unverifiable Internet Text},
  author  = {Lu, Ximing and Acuna, David and Jung, Jaehun and Hu, Jian and
             Zhang, Di and Diao, Shizhe and Zou, Yunheng and
             Zhang, Shaokun and Cui, Brandon and Liu, Mingjie and
             Kim, Hyunwoo and Ammanabrolu, Prithviraj and Kautz, Jan and
             Dong, Yi and Choi, Yejin},
  journal = {arXiv preprint arXiv:2601.22975},
  year    = {2026}
}

@article{guo2026rethinkingmcq,
  title   = {Rethinking Multiple-Choice Questions for {RLVR}: Unlocking
             Potential via Distractor Design},
  author  = {Guo, Xu and Ge, Qiming and Tong, Jian and Chen, Kedi and
             Zhang, Jin and Yang, Xiaogui and Gao, Xuan and Lv, Haijun and
             Lu, Zhihui and Zou, Yicheng and Guo, Qipeng},
  journal = {arXiv preprint arXiv:2603.12826},
  year    = {2026}
}

@article{tang2026positionbias,
  title   = {Do Large Language Models Plan Answer Positions? Position
             Bias in Multiple-Choice Question Generation},
  author  = {Tang, Xuemei and Duan, Xufeng and Cai, Zhenguang G.},
  journal = {arXiv preprint arXiv:2605.01846},
  year    = {2026}
}

@article{cai2025noisy,
  title   = {Reinforcement Learning with Verifiable yet Noisy Rewards under
             Imperfect Verifiers},
  author  = {Cai, Xin-Qiang and Wang, Wei and Liu, Feng and Liu, Tongliang
             and Niu, Gang and Sugiyama, Masashi},
  journal = {arXiv preprint arXiv:2510.00915},
  year    = {2025}
}

@inproceedings{egashira2026delay,
  title     = {Delay, Plateau, or Collapse: Evaluating the Impact of
               Systematic Verification Error on {RLVR}},
  author    = {Egashira, Kazuki and Vero, Mark and Dekoninck, Jasper and
               Dorner, Florian E. and Staab, Robin and Vechev, Martin},
  booktitle = {Conference on Language Modeling (COLM)},
  year      = {2026},
  note      = {arXiv:2605.02909}
}

@inproceedings{huang2025accuracy,
  title     = {From Accuracy to Robustness: A Study of Rule- and
               Model-based Verifiers in Mathematical Reasoning},
  author    = {Huang, Yuzhen and Zeng, Weihao and Zeng, Xingshan and
               Zhu, Qi and He, Junxian},
  booktitle = {Proceedings of the 2026 Conference on Empirical Methods in
               Natural Language Processing (EMNLP)},
  year      = {2026},
  note      = {arXiv:2505.22203}
}

@article{helff2026gaming,
  title   = {{LLMs} Gaming Verifiers: {RLVR} can Lead to Reward Hacking},
  author  = {Helff, Lukas and Delfosse, Quentin and Steinmann, David and
             H{\"a}rle, Ruben and Shindo, Hikaru and
             Schramowski, Patrick and Stammer, Wolfgang and
             Kersting, Kristian and Friedrich, Felix},
  journal = {arXiv preprint arXiv:2604.15149},
  year    = {2026}
}

@article{ray2026fuzzing,
  title   = {Before the Model Learns the Bug: Fuzzing {RLVR} Verifiers},
  author  = {Ray, Jaideep},
  journal = {arXiv preprint arXiv:2606.01066},
  year    = {2026}
}

@article{norman2026reliability,
  title   = {Reliability without Validity: A Systematic, Large-Scale
             Evaluation of {LLM}-as-a-Judge Models Across Agreement,
             Consistency, and Bias},
  author  = {Norman, Justin D. and Rivera, Michael U. and Hughes, D. Alex},
  journal = {arXiv preprint arXiv:2606.19544},
  year    = {2026}
}

@inproceedings{wu2025syntheticaudit,
  title     = {Synthetic Artifact Auditing: Tracing {LLM}-Generated
               Synthetic Data Usage in Downstream Applications},
  author    = {Wu, Yixin and Yang, Ziqing and Shen, Yun and
               Backes, Michael and Zhang, Yang},
  booktitle = {34th USENIX Security Symposium},
  year      = {2025},
  note      = {arXiv:2502.00808}
}

@article{du2023shortcut,
  title   = {Shortcut Learning of Large Language Models in Natural
             Language Understanding},
  author  = {Du, Mengnan and He, Fengxiang and Zou, Na and Tao, Dacheng
             and Hu, Xia},
  journal = {Communications of the ACM},
  volume  = {67},
  number  = {1},
  pages   = {110--120},
  year    = {2023}
}

\appendix
\section{Reproducibility Details}
\label{sec:appendix}
Table~\ref{tab:hyperparams} reports the full training, evaluation, and
compute configuration for Stage 2, including the per-domain sample sizes
matched across the treatment and control arms.
\begin{table}[htbp]
\centering
\small
\setlength{\tabcolsep}{5pt}
\begin{tabular}{ll}
\toprule
Setting & Value \\
\midrule
Base policy model & Qwen3-1.7B \\
Paraphrase model (control construction) & Qwen2.5-3B-Instruct \\
Fine-tuning method & 4-bit NF4 QLoRA, $r{=}16$, $\alpha{=}32$ \\
LoRA target modules & \texttt{q\_proj, k\_proj, v\_proj, o\_proj} \\
RL algorithm & GRPO \citep{shao2024deepseekmath} \\
Training steps per arm & 60 \\
Generations per prompt & 2 \\
Effective batch size & 8 (1 $\times$ 8 accumulation) \\
Max prompt / completion length & 384 / 16 tokens \\
Thinking mode & disabled (\texttt{enable\_thinking=False}) \\
Reward & rule-based exact letter match \\
Train items per domain (matched, both arms) & 1{,}000 \\
Test items per domain & 300 \\
Train/test partition & stable hash on item id, $\approx$80/20 \\
Compute & single 16\,GB GPU (Azure ML) \\
\bottomrule
\end{tabular}
\caption{Training, evaluation, and compute configuration for Phase~2.
Train-set size is matched exactly across the treatment and control arms
to avoid confounding a data-authenticity effect with a data-quantity
effect. All settings are released with the code so the run can be
reproduced or extended to a larger compute budget.}
\label{tab:hyperparams}
\end{table}

\end{document}